\documentclass[a4paper, 10pt, conference]{ieeeconf}      

\IEEEoverridecommandlockouts                              

\usepackage{amsmath}
\usepackage{enumerate}
\usepackage{mathrsfs}
\usepackage{makeidx}         
\usepackage{graphicx}        
\usepackage{multicol}        
\usepackage{subfigure}
\usepackage{epsfig,amssymb,latexsym}
\usepackage{psfrag}

\usepackage{epsfig,amssymb,latexsym}
\usepackage{psfrag}
\usepackage[ruled,vlined]{algorithm2e}

\usepackage{algorithm2e}

\def\QED{~\rule[-1pt]{5pt}{5pt}\par\medskip}

\newcommand{\bi}{\begin{itemize}}
\newcommand{\ei}{\end{itemize}}
\newcommand{\be}{\begin{equation}}
\newcommand{\ee}{\end{equation}}
\newcommand{\bd}{\begin{displaymath}}
\newcommand{\ed}{\end{displaymath}}
\newcommand{\bea}{\begin{eqnarray}}
\newcommand{\eea}{\end{eqnarray}}
\newcommand{\ba}{\begin{array}}
\newcommand{\ea}{\end{array}}
\newcommand{\bc}{\begin{center}}
\newcommand{\ec}{\end{center}}

\newcommand{\nb}{\nonumber}

\newcommand{\re}[1]{(\ref{#1})}

\title{\LARGE \bf
Dynamic analysis of simultaneous adaptation of force, impedance and trajectory}
\author{Y. Li and E. Burdet
\thanks{The authors are with the Department of Bioengineering, Imperial College of Science, Technology and Medicine, London SW72AZ, UK.} 
}

\begin{document}

\maketitle
\thispagestyle{empty}
\pagestyle{empty}

\noindent\emph{When carrying out tasks in contact with the environment, humans are found to concurrently adapt force, impedance and trajectory. Here we develop a robotic model of this mechanism in humans and analyse the underlying dynamics.
We derive a general adaptive controller for the interaction of a robot with an environment solely characterised by its stiffness and damping, using Lyapunov theory.}

\section{System dynamics}
The dynamics of a $n$-degree-of-freedom ($n$-DOF) robot in the operational space are given by
\be
\label{eq.robot}
M(q) \, \ddot{x} + C(q,\dot{q}) \, \dot{x} + G(q) = u + f
\ee
where $x$ is the position of the robot and $q$ the vector of joints angle. $M(q)$ denotes the inertia matrix, $C(q,\dot{q})\dot{x}$ the Coriolis and centrifugal forces, and $G(q)$ the gravitational force, which can be identified using e.g. nonlinear adaptive control \cite{Burdet1998}. $u$ is the control input and $f$ the interaction force.

In \cite{Yang2011}, we have described the control input $u$ in two parts:
\be
u = v + w \,,
\label{e:u}
\ee
with $v$ to track the \emph{reference trajectory} $x_r$ by compensating for the robot's dynamics, i.e.
\be
v = M(q) \, \ddot{x}_e + C(q,\dot{q}) \, \dot{x}_e + G(q) - \Gamma \varepsilon
\ee
where
\be
\dot x_e = \dot x_r-\alpha e \,, \quad e \equiv x - x_r \, , \quad \alpha > 0\, ,
\ee
$\Gamma$ a symmetric positive-definite matrix with minimal eigenvalue $\lambda_{\mbox{min}}(\Gamma)\geqslant \lambda_\Gamma>0$ and
\be
\varepsilon \equiv \dot{e} + \alpha \, e
\ee
the \emph{tracking error}. $w$ is to adapt impedance and force in order to compensate for the unknown interaction dynamics.

\section{Force and impedance adaptation}
Suppose that the interaction force can be expanded as
\bea
f = F^*_{0} + K^*_{S} (x - x^*_0) + K^*_{D} \dot{x} \, ,
\label{e:expansion}
\eea
where the force $F^*_{0}(t)$, stiffness $K^*_{S}(t)$ and damping $K^*_{D}(t)$ are feedforward components of the interaction force, $x^*_0(t)$ is the rest position of the environment visco-elasticity and all of these functions are unknown but periodic with $T$:
\bea
F^*_{0}(t+T) &\equiv& F^*_{0}(t) \, , \quad K^*_{S}(t+T) \equiv K^*_{S}(t) \,,\\
K^*_{D}(t+T)&=&K^*_{D}(t) \, , \quad x^*_0(t+T) = x^*_0(t) \, .
\eea
To simplify the analysis, we rewrite the interaction force as
\be
\label{eq.force}
f \equiv F^* + K^*_{S} \, x + K^*_{D} \, \dot{x}
\ee
where $F^* \equiv F^*_{0}-K^*_{S}x^*_0$ is also periodic with $T$. $w$ in Eq.(\ref{e:u}) is then defined as
\be
w = -F - K_Sx - K_D \dot{x}
\ee
where $K_S$ and $K_D$ are stiffness and damping matrices, respectively, and $F$ is the feedforward force.

By substituting the control input $u$ into Eq.\re{eq.robot}, the closed-loop system dynamics are described by
\be
\label{eq.closed}
M(q) \, \dot{\varepsilon} + C(q,\dot{q}) \, \varepsilon +\Gamma \varepsilon = \widetilde{F} + \widetilde{K}_{S} \, x + \widetilde{K}_{D} \, \dot{x} \, ,
\ee
\bd
\widetilde{F} \equiv F^*-F \,, \quad \widetilde{K}_{S} \equiv K^*_{S}-K_{S}\, , \quad \widetilde{K}_{D} \equiv K^*_{D}-K_{D} \, .
\ed
In this equation, we see that the feedforward force $F$, stiffness $K_{S}$ and damping $K_{D}$ ensure contact stability by compensating for the interaction dynamics. Therefore, the objective of force and impedance adaptation is to minimise these residual errors which can be carried out through minimising the cost function
\bea
\label{eq.jc}
J_c(t) &\equiv& \frac{1}{2}\int^{t}_{t-T} \widetilde F^TQ_F^{-1}\widetilde F+ \mbox{vec}^T(\widetilde K_{S})Q^{-1}_S\mbox{vec}(\widetilde K_{S})\nb\\
&&+\mbox{vec}^T(\widetilde K_{D})Q^{-1}_D\mbox{vec}(\widetilde K_{D}) \, d\tau \, ,
\eea
where $Q_F$, $Q_S$ and $Q_D$ are symmetric positive-definite matrices, and vec$(\cdot)$ stands for the column vectorization operation. This objective is achieved through the following update laws:
{\small
\bea
\label{eq.update-imp1}
\delta F(t) &\equiv& F(t) - F(t-T) \equiv Q_F[\varepsilon(t) -\beta(t) F(t)] \\
\delta K_{S}(t) &\equiv& K_{S}(t) - K_{S}(t-T) = Q_S [\varepsilon(t) x(t)^T - \beta(t) K_{S}(t)]  \nb\\
\delta K_{D}(t) &\equiv& K_{D}(t) - K_{D}(t-T) = Q_D [\varepsilon \, \dot{x}(t)^T - \beta(t) K_{D}(t)] \nb
\eea
}
where $F$, $K_{S}$ and $K_{D}$ are initialised as zero matrices/vectors with proper dimensions for $t\in[0,T)$.

Now that we have dealt with the interaction dynamics, stable trajectory control can be obtained by minimising the cost function
\be
J_e(t) \equiv \frac{1}{2} \varepsilon(t)^T M(q) \, \varepsilon(t) \, .
\ee
Consequently, we use a combined cost function $J_{ce}\equiv J_c + J_e$ that yields concurrent minimisation of tracking error and control effort.

\section{Trajectory Adaptation}
In a typical interaction task, the contact between the robot and the environment is maintained through a desired interaction force $F_d$. Assuming that there exists a desired trajectory $x_d$ yielding $F_d$, i.e. from Eq.(\ref{e:expansion})
\bea
\label{eq.xd}
F_d &=& F^*_{0} + K^*_{S}(x_d-x_0^*) + K^*_{D} \, \dot{x}_d \\
&=& F^* + K^*_{S} \, x_d + K^*_{D} \, \dot{x}_d \, , \quad F^* = F^*_{0} - K^*_{S} \, x_0^* \, , \nb
\eea
we propose to adapt the reference $x_r$ in order to track $x_d$. However, $x_d$ is unknown as the parameters $F^*$, $K^*_{S}$ and $K^*_{D}$ in the interaction force are unknown. Nevertheless, we know that $x_d$ is periodic with $T$ as $F^*$, $K^*_{S}$ and $K^*_{D}$ are periodic with $T$ and we also set $F_d$ to be periodic with $T$.

In the following, we develop an update law to learn the desired trajectory $x_d$. First, we define
\be
\xi_d \equiv K^*_{S} \, x_d+K^*_{D} \, \dot{x}_d \, , \quad \xi_r \equiv K_{S} \, x_r + K_{D} \, \dot{x}_r \, .
\ee
Then, we develop the following update law
\be
\label{eq.update-xir}
\delta \xi_r(t) \equiv \xi_r - \xi_r(t-T) \equiv L^{-T}Q_r (F_d(t) - F(t) - \xi_r(t))
\ee
where $Q_r$ and $L$ are positive-definite constant gain matrices. This update law minimises the error between $\xi_d$ and $\xi_r$, which is described by the following cost function
\bea
\label{eq.jr}
J_r \equiv \frac{1}{2}\int^{t}_{t-T} (\xi_r-\xi_d)^T Q_r^{T}(\xi_r-\xi_d) \, d\tau \, .
\eea
Because of the coupling of adaptation of force and impedance and trajectory adaptation, we modify the adaptation of feedforward force Eq.\re{eq.update-imp1} to
\bea
\label{eq.update-force}
\delta F(t) \equiv Q_F [\varepsilon(t) - \beta(t) F(t) + Q_r^T \delta \xi_r(t)] \, .
\eea
As a result, update laws Eqs.\re{eq.update-xir} and \re{eq.update-force} minimise the overall cost $J=J_c+J_e+J_r$ as shown in Appendix A.

Then, we obtain the update law for trajectory adaptation
\be
\delta x_r \equiv x_r(t) - x_r(t-T)
\ee
by solving
\be
\label{eq.update-xr}
\delta \xi_r = K_S \, \delta x_r + K_D \, \delta \dot{x}_r = K_S\, \delta x_r + K_D \, \frac{d}{dt}(\delta x_r)
\ee
using $\delta \xi_r(t)$ from Eq.\re{eq.update-xir}. According to the convergence of $\delta\xi_r$, $K_S$ and $K_D$ as shown in Appendix A, $x_r$ will converge, as
\be
\delta\xi_r-\xi_d = K_S \delta x_r + K_D \delta\dot{x}_r \, ,
\ee
Upon convergence, the desired interaction force $F_d$ is maintained between the robot and the environment according to Eq.\re{eq.update-xir}. At the same time, the properties with adaptation of force and impedance are preserved which include trajectory tracking and control effort minimisation. However, from the analysis in Appendix A, we cannot draw the conclusion that $F$, $K_S$, $K_D$ and $x_r$ converge to $F^*$, $K_S^*$, $K_D^*$ and $x_d$, respectively, which will require the condition of persistent excitation (PE), similar to classical adaptive control theory \cite{Astrom1995}.

\section{Discussion}\label{sec.remark}
\subsection{No contact}
In a special case when there is \emph{no force applied by the environment} and $F_d$ is also zero, the controller component $w$ will converge to zero. According to the update law Eq.\re{eq.update-xir}, the reference trajectory will not adapt, as expected.

\subsection{No damping}
If we \emph{neglect the damping} component in the interaction force $f$ of Eq.\re{eq.force}, the trajectory adaptation described by Eqs.\re{eq.update-xir} and \re{eq.update-xr} can be simplified to
\bea
\label{eq.update-xr2}
\delta x_r = L^{-T}Q_r (F_d - F - K_{S} \, x_r)
\eea
Correspondingly, the update laws for force and impedance Eq.\re{eq.update-imp1} needs to be modified as
\bea
\label{eq.update-imp2}
\delta F &\equiv& Q_F (\varepsilon - \beta F + Q_r^T\delta x_r) \, ,\\
\delta K_{S} &\equiv& Q_S (\varepsilon \, x^T - \beta K_{S} + x_r^T Q_r^T \delta x_r) \, .\nb
\eea
The stability analysis is similar to the case with damping and is briefly explained in Appendix B.

\subsection{Force sensing}
As in \cite{Yang2011}, force sensing is not required in the proposed framework, in contrast to traditional methods for surface following where the force feedback is used to regulate the interaction force \cite{Jung2001}.

In particular, in a first phase force and impedance adaptation is used to compensate for the interaction force from the environment. During this process, the unknown actual interaction force is estimated when the tracking error $\varepsilon$ goes to zero as can be seen from Eq.\re{eq.closed}: when $\varepsilon=0$, we have
\bea
w=-f.
\eea
Using this estimated interaction force, then a desired force in Eq.\re{eq.xd} can be rendered by adaptation of the reference trajectory $x_r$.

In this sense, it is important to note that \emph{trajectory adaptation should be conducted only when force and impedance adaptation takes effect, which guarantees compensation of the interaction force and tracking of the current reference trajectory}. Nevertheless, as shown in above stability analysis, adaptation of force, impedance and trajectory can be realised simultaneously.

This also suggests that a force sensor should be used if available, as force and impedance adaptation could then be replaced by force feedback. In this way, trajectory adaptation would not depend on the force estimation process and can in principle happen faster than force and impedance adaptation is needed. However, the potential advantages of a force sensor depends on the quality of the signal it could provide, its cost and the complexity of its installation and use.

\section{Appendix}
\subsection{Proof for minimisation of overall cost $J$}\label{app.1}
Considering the definition of $J_r$ in Eq. \re{eq.jr}, we have
{\small
\bea
\label{ineq.jr-1}
&& \delta J_r(t) \equiv J_r(t) - J_r(t-T) \nb \\
&=& \frac{1}{2}\int^{t}_{t-T} [\xi_r(\tau)-\xi_d(\tau)]^T Q_r^{T}[\xi_r(\tau) - \xi_d(\tau)] \, d\tau \nb \\
&&-\frac{1}{2}\int^{t}_{t-T} [\xi_r(\tau) - \xi_d(\tau)]^T Q_r^{T}[\xi_r(\tau-T) - \xi_d(\tau-T)] \, d\tau \nb \\
&&+ \frac{1}{2}\int^{t}_{t-T} [\xi_r(\tau)-\xi_d(\tau)]^T Q_r^{T}[\xi_r(\tau-T)-\xi_d(\tau-T)] \, d\tau \nb \\
&& - \frac{1}{2}\int^{t}_{t-T} [\xi_r(\tau-T)-\xi_d(\tau-T)]^T Q_r^{T} \times \nb \\
&& \quad \quad [\xi_r(\tau-T)-\xi_d(\tau-T)] \, d\tau \nb \\
&=&\frac{1}{2}\int^{t}_{t-T} [\xi_r(\tau)-\xi_d(\tau)]^T Q_r^{T}\delta \xi_r(\tau) \, d\tau \nb \\
&&+ \frac{1}{2}\int^{t}_{t-T} [\xi_r(\tau-T)-\xi_d(\tau-T)]^T Q_r^{T}\delta \xi_r(\tau) \, d\tau \nb \\
&=&\int^{t}_{t-T} [\xi_r-\xi_d-\frac{1}{2}\delta \xi_r]^T Q_r^{T}\delta \xi_r \, d\tau \quad \mbox{(as $\xi_d(t)=\xi_d(t-T)$)}\nb \\
&\leqslant&\int^{t}_{t-T} [Q_r(\xi_r(\tau)-\xi_d(\tau))]^T\delta \xi_r(\tau) \, d\tau \, .
\eea
}
According to Eqs.\re{eq.xd} to \re{eq.update-xir}, we rewrite this inequality as
\bea
\label{ineq.jr}
\delta J_r &\leqslant& \int^{t}_{t-T} [Q_r(\xi_r-F_d+F+\widetilde F)]^T\delta \xi_r \, d\tau \nb\\
&=&\int^{t}_{t-T} (-L^T\delta \xi_r+Q_r\widetilde F)^T\delta \xi_r \, d\tau.
\eea

Consider the difference between $J_c$ of two consecutive periods
\bea
\label{ineq.jc}
&&\delta J_c \equiv J_c-J_c(t-T) \\
&=&\frac{1}{2}\int^{t}_{t-T} [(\widetilde F^TQ_F^{-1}\widetilde F-\widetilde F^T(\tau-T)Q_F^{-1}\widetilde F(\tau-T))\nb\\
&&+ \mbox{tr}(\widetilde K^T_{S}Q^{-1}_S\widetilde K_{S}-\widetilde K^T_{S}(\tau-T)Q^{-1}_S\widetilde K_{S}(\tau-T)\nb\\
&&+ (\widetilde K^T_{D}Q^{-1}_D\widetilde K_{D}-\widetilde K^T_{D}(\tau-T)Q^{-1}_D\widetilde K_{D}(\tau-T))] \, d\tau \nb
\eea
where tr$(\cdot)$ stands for the trace of a matrix. We consider that
\bea
\label{ineq.F}
&&\widetilde F^T(\tau) Q_F^{-1}\widetilde F(\tau) - \widetilde F^T(\tau-T)Q_F^{-1}\widetilde F(\tau-T) \nb \\
&=&[\widetilde F^T(\tau) Q_F^{-1}\widetilde F(\tau) - \widetilde F^T(\tau) Q_F^{-1} \widetilde F(\tau-T)] \nb \\
&& + [\widetilde F^T(\tau) Q_F^{-1}\widetilde F(\tau-T)-\widetilde F^T(\tau-T) Q_F^{-1} \widetilde F(\tau-T)] \nb \\
&=& -\widetilde F^T(\tau) Q_F^{-1}\delta F(\tau) - \widetilde F^T(\tau-T) Q_F^{-1} \delta F(\tau) \nb \\
&=& -(2\widetilde F^T(\tau) + \delta F(\tau)) Q_F^{-1} \delta F(\tau) \nb \\
&\leqslant& -2\widetilde F^T(\tau) Q_F^{-1} \delta F(\tau) \nb \\
&=& -2\widetilde F^T(\tau) [\varepsilon(\tau) - \beta(\tau) F(\tau) + Q_r^T\delta \xi_r(\tau)]
\eea
Then, similarly, we have
\bea
\label{ineq.sd}
&&\mbox{tr}[\widetilde K^T_{S}(\tau) Q^{-1}_S \widetilde K_{S}(\tau) - \widetilde K^T_{S}(\tau)(\tau-T)Q^{-1}_S\widetilde K_{S}(\tau-T)]\nb\\
&&\leqslant-2\mbox{tr}\{\widetilde K^T_{S}(\tau)[\varepsilon(\tau)x^T(\tau) - \beta(\tau) K_{S}(\tau)]\} \nb \\
&&\mbox{tr}[\widetilde K^T_{D}(\tau) Q^{-1}_d \widetilde K_{D}(\tau) - \widetilde K^T_{D}(\tau-T)Q^{-1}_D\widetilde K_{D}(\tau-T)] \nb \\
&&\leqslant -2\mbox{tr}[\widetilde K^T_{D}(\tau)(\varepsilon(\tau) \dot{x}^T(\tau) - \beta(\tau) K_{D}(\tau))]
\eea
Substituting Ineqs. \re{ineq.F} and \re{ineq.sd} into Eq.\re{ineq.jc} and considering Ineq. \re{ineq.jr} yields
\bea
\label{ineq.jr-jc}
&&\delta J_r + \delta J_c \leqslant \int^{t}_{t-T} \!\!\!\! -\delta\xi_r^T L \delta\xi_r- \widetilde F^T(\varepsilon-\beta F)  \\
&& - \, \mbox{tr}[\widetilde K^T_{S}(\varepsilon x^T-\beta K_{S})]
- \mbox{tr}[\widetilde K^T_{D}(\varepsilon \dot{x}^T - \beta K_{D})] \, d\tau \nb \,.
\eea
The rest is to deal with the residual in the above inequality, which is similar to that in \cite{Yang2011}. For completeness, we show the outline in the following. In particular, we consider the time derivative of $J_e$
\bea
\dot J_e&=&\varepsilon^TM(q,\dot q)\dot \varepsilon+\frac{1}{2}\varepsilon^T\dot M(q,\dot q)\varepsilon\nb\\
&=&\varepsilon^TM(q,\dot q)\dot \varepsilon+\frac{1}{2}\varepsilon^TC(q)\varepsilon
\eea
as \cite{TheoryRobotControl}
\be
z^T \dot{M} z \equiv z^T C z \quad \forall z \, .
\ee
Considering the closed-loop dynamics Eq.\re{eq.closed}, above equation can be written as
\bea
\dot{J}_e(t) \equiv \varepsilon^T (\widetilde{F}^T + \widetilde{K}^T_{S} \, x + \widetilde K^T_{D} \, \dot{x} - \Gamma \varepsilon) \, .
\eea
Integrating $\dot{J}_e$ from $t-T$ to $t$ and considering Ineq. \re{ineq.jr-jc}, we obtain
\bea
\delta J& =& \delta J_c+\delta J_r+\delta J_e \nb\\
& \leqslant& \int^{t}_{t-T}  \!\!\!\!\!\!  -\varepsilon^T \Gamma \varepsilon-\delta\xi_r^T L \delta\xi_r\nb\\
&&+\beta[ \widetilde{F}^T F + \mbox{tr}(\widetilde{K}^T_{S} K_{S} + \widetilde{K}^T_{D} K_{D})] \, d\tau\nb\\
&=& \int^{t}_{t-T}  \!\!\!\!\!\!  -\varepsilon^T \Gamma \varepsilon-\delta\xi_r^T L \delta\xi_r\nb\\
&&-\beta[ \widetilde{F}^T \widetilde{F} + \mbox{tr}(\widetilde{K}^T_{S} \widetilde{K}_{S} + \widetilde{K}^T_{D} \widetilde{K}_{D})]\nb\\
&&+\beta[ \widetilde{F}^T F^* + \mbox{tr}(\widetilde{K}^T_{S} K^*_{S} + \widetilde{K}^T_{D} K^*_{D})] \, d\tau \, .
\eea
A sufficient condition for $\delta J\leqslant 0$ is
\bea
&&\lambda_\Gamma \|\varepsilon\|^2+\lambda_L\|\delta\xi_r\|^2 +\beta(\|\widetilde{F}\|^2 + \|\widetilde{K}_{S}\|^2 + \|\widetilde{K}_{D}\|^2)\nb\\
&&-\beta( \|\widetilde{F}\| \|F^*\| + \|\widetilde{K}_{S}\| \|K^*_{S}\| + \|\widetilde{K}_{D}\| \|K^*_{D}\|) \geq 0 \, .
\eea
where $\lambda_\Gamma$ and $\lambda_L$ are the minimal eigenvalues of $\Gamma$ and $L$, respectively. Therefore, $\|\varepsilon\|$, $\|\delta\xi_r\|$, $\|\widetilde F\|$, $\|\widetilde K_S\|$ and $\|\widetilde K_D\|$ are bounded. In particular, they satisfy
\bea
&&\lambda_\Gamma \|\varepsilon\|^2+\lambda_L\|\delta\xi_r\|^2 +\frac{\beta}{2}(\|\widetilde{F}\|^2 + \|\widetilde{K}_{S}\|^2 + \|\widetilde{K}_{D}\|^2)\nb\\
&&\leq\frac{\beta}{2}( \|F^*\|^2 + \|K^*_{S}\|^2 + \|K^*_{D}\|^2) \, .
\eea
By choosing large $\lambda_\Gamma$ and $\lambda_L$, $\|\varepsilon\|$ and $\|\delta\xi_r\|$ can be made small.

\subsection{Proof for minimisation of overall cost when neglecting damping}\label{app.2}
Consider the cost function
\bea
J_r' \equiv \frac{1}{2}\int^{t}_{t-T} (x_r - x_d)^T K_S^{*T} Q_r^{T} (x_r - x_d) \, d\tau \, .
\eea
Following similar procedures to Ineqs. \re{ineq.jr-1}, \re{ineq.jr}, we obtain
\bea
\label{ineq.jr'}
\delta J_r'\leqslant \int^{t}_{t-T} [-L^T\delta x_r+ Q_r(\widetilde F+\widetilde K_S x_r)]^T\delta x_r \, d\tau
\eea
Considering further the cost function
\be
J_c' \equiv \frac{1}{2}\int^{t}_{t-T} \! \! \widetilde{F}^TQ_F^{-1} \widetilde{F} +\mbox{vec}^T(\widetilde{K}_{S})Q^{-1}_S\mbox{vec}(\widetilde{K}_{S}) \, d\tau \, .
\ee
and following similar procedures from Ineqs.\re{ineq.jc} to \re{ineq.jr-jc}, we obtain
{\small
\bea
\label{ineq.jr'-jc'}
&&\hspace{-5mm}\delta J_r'+\delta J_c'\\
&&\hspace{-5mm}\leqslant \int^{t}_{t-T} \!\!\!\!\!\! -\delta x_r^T L\delta x_r-\widetilde F^T(\varepsilon-\beta F) -\mbox{tr}[\widetilde K^T_{S}(\varepsilon \, x^T-\beta K_{S})] \,  d\tau \, .\nb
\eea
}
The rest is similar to the case with damping and thus omitted.




\end{document}